\pdfoutput=1

\documentclass[11pt]{article}

\usepackage[]{acl}

\usepackage{times}
\usepackage{latexsym}
\usepackage{multirow}

\usepackage[T1]{fontenc}

\usepackage[utf8]{inputenc}

\usepackage{microtype}

\usepackage{graphicx}
\usepackage{xcolor}
\definecolor{ao(english)}{rgb}{0.0, 0.5, 0.0}

\newcommand{\win}[1]{{\colorbox{cyan}{\sf \color{white}{#1}}}}

\title{Am I Me or You?\\  State-of-the-Art Dialogue Models Cannot Maintain an Identity}

\author{Kurt Shuster \quad Jack Urbanek \quad Arthur Szlam \quad Jason Weston\\
\\
 Facebook AI Research
}

\begin{document}
\maketitle

\begin{abstract}
State-of-the-art dialogue
models still often stumble with regards to factual accuracy and self-contradiction.  
Anecdotally, they have been observed to fail to maintain character identity throughout discourse; and more specifically, may take on the role of their interlocutor.
In this work we formalize and quantify this deficiency,
and show experimentally through human evaluations that this is indeed a problem.  In contrast, we show that discriminative models trained specifically to recognize who is speaking can perform well; and further, these can be used as automated metrics.  
Finally, we evaluate a wide variety of mitigation methods, including changes to model architecture, training protocol, and decoding strategy.  Our best models reduce mistaken identity issues 
by nearly 65\% according to human annotators, while simultaneously improving engagingness. 
 Despite these results, we find that
 maintaining character identity still remains a challenging problem.
\end{abstract}

\section{Introduction}


The exchange of stories from one's past, or descriptions of activities in one's present, are a fundamental part of human discourse.  Trustworthy human conversationalists keep their stories roughly straight within a conversation.  An interlocutor taking on {\it your} own stories and persona as {\it theirs} is especially jarring and unnatural.   However, despite the improvements in state-of-the-art open-domain dialogue modeling, both in terms of distributional accuracy metrics like perplexity, and subjectively in terms of human judgements \citep{adiwardana2020meena, roller2020recipes}, interactions with those agents reveal that they cannot keep their stories straight.   In particular, they are likely to take on the role of their interlocutor;  for example, if an agent's partner says they are a software engineer, the agent is likely to say it is a software engineer too \citep{roller2020recipes}, or worse, appropriate their partners just told tale of a trip to NAACL as their own. Some example failure cases are given in Table \ref{tab:example_misidentity}, where models incorrectly take on the name, role or activities of their partner instead of their assigned role.
These failures are related to the general problems of repetition in language models \citep{holtzman2019curious}, the weak influence of word order \citep{sinha2021masked} and inability to avoid contradictions \citep{nie-etal-2021-like}.



\begin{table}[t!]
    \centering
    \resizebox{\linewidth}{!}{
    \begin{tabular}{|ll|}
    \hline
    \multicolumn{2}{|c|}{\textit{\textbf{Setting}: Barn, Farm} } \\
    \multicolumn{2}{|c|}{\textit{\textbf{Partner}: {\it \color{red}{Thief}}} } \\
    \multicolumn{2}{|c|}{\textit{\textbf{Self}: {\it \color{ao(english)}{Guest}}}} \\
    \hline
    \multicolumn{2}{|l|}{\textbf{Guest}: Hail who goes there?! } \\
    \multicolumn{2}{|l|}{\textbf{Thief}: Hello friend I mean you no harm}\\
    \hline
    \multicolumn{2}{|l|}{\textbf{\textit{Model Response}} } \\
    \multicolumn{2}{|l|}{\textbf{Guest}: \textit{I am not your friend, I am your {\bf \color{red}{thief}!} I am here}} \\
    \multicolumn{2}{|l|}{\textit{to steal from you!}} \\
    \hline
    \multicolumn{2}{|l|}{\textbf{Character Token Probabilities}}\\
    {\bf \color{red}{Thief}}: 0.39 & {\bf \color{ao(english)}{Guest}}: 0.23 \\
    \hline
    \hline
    \multicolumn{2}{|c|}{\textit{\textbf{Setting}: Castle Kitchens, Inside Palace} } \\
    \multicolumn{2}{|c|}{\textit{\textbf{Partner}: {\it \color{red}{Hunter}}} } \\
    \multicolumn{2}{|c|}{\textit{\textbf{Self}: {\it \color{ao(english)}{Helper}}}} \\
    \hline
    \multicolumn{2}{|l|}{\textbf{Hunter}: Hi } \\
    \multicolumn{2}{|l|}{\textbf{Helper}: Hello. What brings you to the kitchen?}\\
    \multicolumn{2}{|l|}{\textbf{Hunter}: I am chopping some meat that is to be sold. } \\
    \multicolumn{2}{|l|}{\textbf{Helper}: Could you spare me a piece? I'm starving.}\\
    \multicolumn{2}{|l|}{\textbf{Hunter}: It's 3 cents a pound, if you have 3 cents then yeah} \\
    \hline
    \multicolumn{2}{|l|}{\textbf{\textit{Model Response}} } \\
    \multicolumn{2}{|l|}{\textbf{Helper}: \textit{I don't have any money. I'm just a poor {\bf \color{red}{hunter}}}} \\
    \multicolumn{2}{|l|}{\textit{trying to make a living.}} \\
    \hline
    \multicolumn{2}{|l|}{\textbf{Character Token Probabilities}}\\
    {\bf \color{red}{Hunter}}: 0.39 & {\bf \color{ao(english)}{Helper}}: 0.12 \\
    \hline
    \end{tabular}
}
    \caption{Failures of a 2.7B parameter transformer pre-trained on 1.5B examples from a large dialogue corpora \citep{roller2020recipes} and fine-tuned on LIGHT \citep{urbanek2019learning}. 
    The bold words in {\bf \color{red}{red}} highlight the model mistaking its identity for its partner's. (\textbf{Top}) The model believes it is a {\bf \color{red}{thief}}, rather than a {\bf \color{ao(english)}{guest}}. (\textbf{Bottom}) the model believes it is a {\bf \color{red}{hunter}} rather than a  {\bf \color{ao(english)}{helper}}. Token probabilities are given at the position of the mistake for the two names.
    }
    \label{tab:example_misidentity}
\end{table}


In this work we formalize and quantify this behavior, show that to some extent it can be detected automatically with a specifically trained classifier, and then study a wide variety of mitigations. 
These include
multi-objective training, 
unlikelihood training \citep{welleck2019neuraltext},
classifier-assisted re-ranking based generation, 
and several forms modifying the attention mechanisms of the decoder in a sequence to sequence model. 
Our best methods can reduce mistaken identity issues by 65\%, while simultaneously improving in-conversation engagingness; indeed, our models that can stick to their role in conversation are judged by humans to be significantly more engaging than their baseline counterparts. Despite these advances, we find that there is still considerable space to improve these results further in future work.

We make publicly available both our trained models and code to reproduce results\footnote{\scriptsize{\url{https://parl.ai/projects/light_whoami/}}}.

\section{Related Work}

\paragraph{Role-Playing in Open-Domain Dialogue} Much recent work has explored training open-domain dialogue models on large and small dialogue corpora, with the former imbuing raw conversational ability and the latter providing necessary conversational skills. Most crowd-sourced datasets require acting out a role to some capacity in conversation (though indeed \citet{mazare2018trainingmillions} study extraction of roles from raw data). Some involve providing persona lines that a model must assume throughout the conversation \citep{zhang2018personalizing,dinan2019second,xu2021goldfish}; others require more subtle "roles", such as a listener \citep{rashkin2019empathetic}, or a teacher and student \citep{dinan2018wizard,gopalakrishnan2019topical,zhou2018dataset,komeili2021internetaugmented}. 

\paragraph{Consistency in Open-Domain Dialogue} A common paradigm in the state of the art of open-domain dialogue involves concatenating all relevant contextual information as input to a sequence to sequence neural model (e.g., transformers \citep{vaswani2017attention}) to obtain a conditioned response. Such models can yield human-like and engaging responses \citep{adiwardana2020meena,roller2020recipes}. Nevertheless, 
various consistency issues still
plague such models. Recent studies have indicated that hallucination of incorrect knowledge is still far from a solved issue \citep{shuster2021retrieval, santhanam2021rome}, with some proposing specific datasets and tools for measuring precisely the levels of this undesired attribute \citep{liu2021tokenlevel}. Another clear example of failure is the short-term memory of state-of-the-art models \citep{xu2021goldfish}, sometimes due to the lack of long-form training data or long-context models but often due to simply the modeling itself. 

To address consistency issues, a variety of methods have been explored.
In the context of knowledge-grounded dialogue, different
ways to attend most effectively over provided contextual information have been explored \citep{10.1145/3357384.3357889,2020yeknowledgegrounded,prabhumoye-etal-2021-focused,wang2019improving}. These works find that considering factual documents separately (in some capacity) improves model grounding. We explore such methods, but in the context of character identity. 

Another general problem 
is that of contradictions.  \citet{nie-etal-2021-like} collect a dataset of contradictions in dialogue, and train classifiers that help re-rank model outputs at inference time; \citet{li2019dontsaythat} explore unlikelihood training \citep{welleck2019neuraltext} to reduce repetition and contradiction, among other undesired traits, in model generations. 
The character identity issue we study in this work can be seen as an important class of contradictions, but to the 
 best of our knowledge, has not been explicitly focused on.

\section{Methods}


\subsection{Problem Setting}
\label{sec:prelims}

We consider a two-party chat setting. The context provided to a model includes: (i) the name of its character and the partner's character;
(ii) an extended description of its own character; (iii) and, information about the area in which the conversation takes place. The responsibility of the model is to engage its conversational partner, with no other goal prescribed; however it should stay within character and within the bounds of the defined setting.

We operate in the context of LIGHT \citep{urbanek2019learning}, consisting of grounded fantasy role-playing game conversations. The LIGHT environment involves humans and models interacting with thousands of objects in hundreds of locations, all while assuming the roles of one of hundreds of characters. 
The dataset consists of roughly 8.5k dialogues spanning 111k utterances. 
It is an ideal setting for this study because of the rich and varied personas with explicit backstories.

To quantify the character identity problem, we take a state-of-the-art dialogue agent (specifically,  BlenderBot \citep{roller2020recipes}) fine-tuned on the LIGHT dialogue dataset and ask human annotators if the agent mistakes its identity based on its utterances in context.  The agent conditions its response on the LIGHT context 
and prior utterances in the dialogue history.  We see in Table \ref{tab:full_char_classification_test} that in roughly $6.5$ percent of utterances the model mistakes its identity; this corresponds to a mistake in approximately $35$ percent of conversations.

BlenderBot uses a Byte-Level BPE tokenizer \citep{radford2019language}; an artifact from the BlenderBot pre-training is that it only considers 128 such tokens in the past, and thus has no mechanism for recovering truncated information about the LIGHT context in later conversational turns. Our second baseline lengthens the input context to 1024 BPE tokens, which allows the entire context for \textit{every example} to fit into the truncation length of the model; we follow methods employed in \citet{xu2021goldfish} to extend the positional embeddings of the model.  We see in Table \ref{tab:full_char_classification_test} that this actually makes the problem worse, resulting in $7.4$ percent of utterances with mistaken identity (corresponding to a failure in approximately $38$ percent of conversations).

\subsection{Measuring Role-Playing Accuracy: RPA}
\label{sec:ica}

We first define a metric, role-playing accuracy (RPA), to denote how often a model's responses are ``in-character''; by this, we mean how often the model's response could feasibly be said by their character, given their assigned character identity. 
Measuring RPA is a non-trivial task for a variety of reasons. First, some conversations involve pairs that can reasonably say similar things (priest vs. priestess, man vs. woman, wizard vs. witch). Second, opening lines are often more generic (``hello'', ``how fare your travels today''), so \textit{either} character can say it in conversation. The third reason stems from the data that we study; we are relying on crowdsourced data in which humans are required to portray their characters. Some crowdworkers may be better than others, and there may be some noise in the dataset in which, e.g., a horse may proclaim its love for a queen, or a knight may discuss at length the kingdom's tax collecting.

We thus train models specifically designed to identify whether a candidate response from a model fits the model's role. We employ poly-encoder transformers \citep{humeau2019polyencoder} to learn this metric, and structure the task as a ranking one; the model receives the LIGHT setting and prior utterances of dialogue as input, as well as the response currently under consideration, and the model must choose the correct character from a fixed set of candidates. 

We also explore RPA classifiers trained on all partially complete sequences of labels, such that the classifiers can determine the character speaking without requiring the full utterance; we call these left-to-right (LTR) RPA classifiers.  Further details about
 how our RPA classifiers are built is 
 given in Appendix \ref{sec:rpa_train_data_appendix}.

\subsection{Mitigations}
In this section we describe several strategies for improving the role-playing accuracy of dialogue agents, specifically ways to improve our transformer baselines. 

\subsubsection{Re-ranking Model Outputs via RPA}
\label{sec:methods_reranking}

We can employ an RPA classifier in response generation by using it to rank candidate model outputs. 
\paragraph{Utterance Re-Ranking:}
Given a set of candidate responses, the RPA classifier can re-score the set and return the response yielding the highest probability of staying in character (according to the RPA score on the complete candidate generations). 
The dialogue models employ beam-search to generate responses, and the candidates for re-ranking are the beams within beam-search. We also try nucleus sampling \citep{holtzman2019curious} and delayed beam-search \citep{massarelli2019decoding} to see whether more diverse candidates have any affect. 
\paragraph{Partial And Complete Efficient Re-ranking (PACER):} 
Re-ranking only the final beam candidates may well be suboptimal because it is well known that those candidates are not very diverse \citep{kulikov2018importance}, meaning there may not be any good candidates to choose from in this final set. 
In order to generate utterances that agree with our classifiers, a possible improvement is to generate the utterance such that partial generations also agree with the classifier when generating left-to-right, ensuring that good candidates are surfaced.
With access to LTR RPA classifiers, we can apply re-ranking to partial sequences.

Unfortunately, re-ranking at every step of beam search, for every token requires significant computation, such as in the recent FUDGE method \citep{yang-klein-2021-fudge}. FUDGE re-scores tokens at each decoding step by multiplying the classifier probability with each token probability, and renormalizing, 
which is used for control tasks with lightweight classifiers in order to be tractable. 

In our proposed approach, called PACER, we re-score candidate tokens, for each beam, according to the probability that their inclusion yields the appropriate character classification, and then finally re-rank the complete candidate beams. To make this efficient, we crucially score only a small proportion of decoding steps (e.g., 5\% of token positions) as well as for only a few candidate re-scored tokens (e.g., top 10 only). We can control these hyperparameters to explore the speed vs. accuracy trade-off.

\subsubsection{Unlikelihood}
\label{sec:methods_unlikelihood}

We explore utilizing an unlikelihood (UL) loss \citep{welleck2019neuraltext} to force the model to stay in character during training. Unlikelihood training works as a counter to the standard maximum likelihood (MLE) training of language models; while MLE training pushes the model to generate the correct tokens, UL training pushes the model to \textit{not} generate \textit{incorrect} tokens. 

While training on the LIGHT dataset with standard NLL loss, with some fixed probability we consider a candidate model generation for UL loss. The full generation is sent to the RPA classifier; if the generation is classified as coming from the incorrect character, we examine each partial generated sequence of the output, and send these sequences to the LTR RPA classifier to determine whether the candidate partial sequences match the model's character.  We apply UL loss to tokens that yield the \textit{wrong} character classification.

\subsubsection{Multi-objective Training}
\label{sec:methods_multiobjective}

The RPA classifiers utilize the LIGHT setting and prior utterances of dialogue history to determine which character generates a candidate response. We hypothesize that the generation models themselves should be able to pick out and utilize these components as well. However, the RPA classifier models are trained explicitly for this task, whereas the seq2seq models are trained only to generate a plausible continuation of a dialogue history.

We thus explore a setup in which the generation models are trained to identify the speaker of an utterance as well. To do this, we use the output representations from the model (either encoder + decoder, or decoder only) as inputs to $n_{MO}$ additional transformer layers, where we vary $n_{MO} \in \{0, 2\}$. The final outputs are used to compute a character score, similarly to the RPA classifier. 

The model can then be trained piece-wise. After initializing the model weights with those trained on the LIGHT response generation task, we then train \textit{only} the extra layers with \textit{only} the character classification objective; once the classifier achieves suitable performance on the task, we can begin to back-propagate the character classification objective multi-tasking with the dialogue task itself to the generation model directly,
in the hope that the model learns to update its internal representations of the context and/or the decoded response.

\subsubsection{Expanded Decoder Attention}
\label{sec:methods_expanded_attention}

Maintaining identity relies on the model's capacity to understand which inputs from the conversational history are pertinent when generating a continuation of the preceding dialogue. In a standard, open-domain chit-chat scenario, the model has free reign to decide which elements of the context it would like to condition on when generating a response, as we are dealing with a nearly unconstrained output space (so long as the output follows plausibly from the input). In LIGHT, however, we want to emphasize certain components of the context more so than others; specifically, when role-playing as a character, we want the model to always be reminded of its role, so that it can conditionally generate an optimal response \textbf{while staying in character}. In this lens, one can view the task as
``grounding" on one's character information when conversing.

\paragraph{Profile Grounding}
Inspired by models demonstrating good performance in knowledge-grounded dialogue \citep{10.1145/3357384.3357889,2020yeknowledgegrounded,prabhumoye-etal-2021-focused,wang2019improving}, we propose a simple extension to the transformer seq2seq architecture, specifically the decoder, to ensure the model knows to condition on the profile. The standard transformer decoder first uses self-attention over the decoded response, and then cross-attention over the encoder outputs. We add a third attention step, \textit{expanded attention}, that attends \textit{again} over an extracted \textit{subset} of the input context (encoded separately from the normal context). We explore various subsets of the context to determine which are most important for both RPA and other automated metrics, and call this method ``Profile'' grounding as the subsets generally include the character and role description. We utilize the \textit{exact same} (shared) parameters for both the normal cross-attention and the expanded attention; thus, model size is not affected.

\paragraph{Automated Grounding}
Instead of directly telling the model what to re-attend to, we also explore whether the model can learn to do this automatically, based on its own (or other) representations of the context.
The first method we consider is examining the {\bf decoder attention weights}. Specifically, we use the attention weights from the decoder over the {\em full context} to choose $k$ tokens to re-attend to. This operation is done on a per-layer basis, and thus allows different decoder layers to re-attend to (potentially different) components of the input.

The second method we consider is a {\bf trainable mask}; this involves feeding the encoded context through a ``mask'' layer to select various tokens to re-attend to. Specifically, we feed the context through a linear projection layer followed by a softmax to select the top-$k$ tokens. This set of tokens is then re-encoded by the encoder and fed to the decoder as the expanded attention context.

Finally, we explore using the {\bf classifier attention weights} over the context \textit{from the RPA classifier itself}. Intuitively, the RPA classifier has learned what components of the input are necessary for determining which character is speaking; if we look at these attention weights when considering the model's character, we know what the classifier thinks is important to use.


\paragraph{Combined Methods}
We also consider combining expanded attention with re-ranking methods,  or with  multi-objective training, to  see if the combination can improve results. For the latter we use the automated grounding trainable mask method. 

\section{Experimental Results}

\begin{table}
\centering
\resizebox{\linewidth}{!}{
\begin{tabular}{l|rr|rr}
\hline
\# Prior & \multicolumn{2}{c}{No LIGHT Context} & \multicolumn{2}{c}{LIGHT Context} \\
Utterances & {H@1/427} & H@1/2 & H@1/427 & H@1/2 \\
\hline
0 & 10.4 & 60.3 & 77.6 & 77.7 \\
4 & 87.3 & 87.4 & 86.5 & 86.5\\
All & 85.7 & 86.7 & 89.3 & 89.8 \\
\end{tabular}
}
\caption{\label{tab:character_classifiers_valid}
RPA classifier performance on the \textbf{validation} set, as measured by Hits@1/427 and Hits@1/2 (all characters and participant characters as candidates, respectively). Each model is trained and evaluated with that \# of prior utterances.
}
\vspace{3mm}
\scriptsize
\centering
\resizebox{\linewidth}{!}{
\begin{tabular}{l|rrr|rrr}
\hline
\textbf{Train} & \multicolumn{6}{c}{\textbf{Ranking Accuracy (Hits@1/427)}}\\
\textbf{split}& \multicolumn{6}{c}{\# Eval Contextual Utterances} \\
\hline
& \multicolumn{3}{c|}{\textbf{Full Datasplit}} & \multicolumn{3}{c}{\textbf{LTR Datasplit}}\\
& 0 & 4 & All & 0 & 4 & All \\
\hline
LTR & 64.8 & 84.3 & 83.9 & 61.7 & 80.5 & 80.5 \\
Full & 31.0 & 86.5 & 84.9 & 27.8 & 75.3 & 74.9 \\
\hline
\end{tabular}
}
\caption{\label{tab:left_to_right_classifier_valid}
RPA classifier performance on the \textbf{validation} set, comparing a partial-sequence trained model (``LTR'') to one trained only on full sequences (``Full''). Models were trained with 4 prior utterances of context.
}
\end{table}

\begin{table*}
\small
\centering
\begin{tabular}{l|rrr|rrrr}
\hline
               & \multicolumn{3}{c}{Automatic Metrics } & \multicolumn{4}{c}{Human Evaluations }\\
\textbf{Model} & PPL$\downarrow$ & F1$\uparrow$ & RPA$\uparrow$ & Mistaken & All-Good$\uparrow$  & Mis. Id. & Engaging$\uparrow$\\
& & & &  Identity$\downarrow$ & & in Conv.$\downarrow$ &\\
\hline
Human & - & - & 92.68 & 1.34\%\\
\hline
\multicolumn{6}{l}{\textbf{\textit{Baselines}}} \\
128-Truncate Vanilla Baseline & 12.64 & 15.69 & 87.61 & 6.45\% & 76.0\% & 35.1\% & 4.04\\
1024-Truncate Vanilla Baseline & 12.43 & 15.68 & 87.71 & 7.35\% & 75.0\%  & 38.4\% & 4.16\\
\hline
\multicolumn{6}{l}{\textbf{\textit{Re-rankers}}} \\
128-Truncate Baseline + RPA Re-Ranker & - & 15.87 & 92.09 & 5.56\% & 80.3\% & 34.7\% & 4.14\\
128-Truncate Baseline + PACER & - & 15.85 & 92.78 & 4.27\% & 73.9\% & 33.7\% &  3.96\\
\hline
\multicolumn{6}{l}{\textbf{\textit{Modified Training Objectives}}} \\
RPA Unlikelihood (Top-1 Token) & 13.10 & 15.18 & 87.48 & 7.13\% & 72.8\% & 39.4\% &  3.87 \\
RPA Unlikelihood (All Tokens) &  13.31 & 14.77 & 88.07 & 10.51\% & 67.7\% & 43.0\% &  3.87 \\
Multi-Objective (Vanilla, Dec. Only)  &  12.86 & 15.67 & 87.67 & 10.00\% & 74.8\% & 49.0\% & 4.21\\
\hline
\multicolumn{6}{l}{\textbf{\textit{Expanded Attention Methods}}}\\
Profile (128, 2 rounds over ABC) & 12.37 & 15.74 & 91.70 & 4.82\% & 81.6\% & 28.4\% & 4.18\\
Profile (1024, 2 rounds over ABCD) & \textbf{12.23} & 15.66 & 92.18 & 4.00\% & 83.8\% & 23.8\% & \textbf{4.34}\\
Automated (1024, Classifier Attn) & 12.27 & 15.75 & 90.93 & 5.51\% & 76.0\% & 29.1\% & 4.04\\
Automated + MO (1024, Dec. Only) & 13.01 & 15.52 & 88.95 & 4.43\% & 78.6\% & 23.0\% &  4.12 \\
\hline
\multicolumn{6}{l}{\textbf{\textit{Expanded Attention + Re-ranker Methods}}}\\
Profile (128) + RPA Re-ranker & - & \textbf{15.88} & 95.16 & \textbf{2.23\%} & 84.4\% & \textbf{14.7\%} & 4.24 \\
Profile (128) + PACER & - & 15.79 & \textbf{95.31} & 4.07\% & \textbf{85.7\%} & 24.5\% & 4.32 \\
\hline
\end{tabular}
\caption{\label{tab:full_char_classification_test}
Automated metrics and human evaluations for various models considered throughout the paper on the LIGHT \textbf{test} set. RPA (Role-Playing Accuracy) is measured by the 4-utterance LTR classifier, see Sec. \ref{sec:ica}.  The  human evaluations are per utterance, except for Engaging and Mistaken Identity in Conversation (with the latter indicating \% of conversations with mistaken identity).
}
\end{table*}

\subsection{RPA Classifiers}

We first assess the quality of our RPA classifiers.  We experiment with either 0, 4, or All prior context utterances, 
for both the standard and left-to-right (LTR) models, and either using the LIGHT context or not. We measure hits@1/427, where the model must correctly identify the character speaking out of 427 characters from the validation set, or hits@1/2 between the two speakers.

%
Results are given in Table \ref{tab:character_classifiers_valid}, where each model is evaluated on the datasplit on which it is trained; Table \ref{tab:character_classifiers_valid_full} in the Appendix shows results across all splits for all models, in which we see that each model performs best on the split on which it is trained when compared to the others, and that our best models can be relatively successful on this task. 
Including the LIGHT context improves RPA performance nearly across the board; with no context and no utterances of dialogue history, the models can only accurately identify the character less than 20\% of the time. 

Results for the LTR RPA classifiers are in Table \ref{tab:left_to_right_classifier_valid}. We experiment with a 4-utterance LTR model; the LTR classifiers perform nearly as well as the full classifiers on the full datasplit, and outperform them quite handily on the LTR split.  Given the robustness of the LTR RPA classifiers, we use this model for computing RPA throughout the remaining results, unless otherwise specified.

\subsection{Baseline Generation Performance}

We next train baseline models for the dialogue generation task itself.
Performance on the LIGHT dataset test split 
for our baseline models can be found in Table \ref{tab:full_char_classification_test}, with results on the validation set in Table \ref{tab:full_char_classification_valid} in the Appendix. While lengthening the context from 128 to 1024 tokens yields better perplexity, the model obtains worse F1 scores and does not improve much if at all on role playing ability, both when measured by the RPA classifiers and in human evaluations (see also Table \ref{tab:human_evals}). Further detailed training and optimization specifications are given in Appendix \ref{sec:training_details}.

\begin{table}
\centering
\resizebox{\linewidth}{!}{
\begin{tabular}{lll|rrrr}
\hline
\textbf{Re-ranker} & \multicolumn{2}{c}{Params} & F1 & RPA & Cost\\
& \# Toks & Freq. \\
\hline
None  &  0  &  0  &  15.8  &  88.4  &    1x \\
Complete-only
&  0  &  0  &  16.0  &  93.0  &   1.1x \\
Partial-only  &  10  &  5\%  &  15.9  &  88.6  &   1.3x \\
Partial-only &  10  &  33\%  &  158  &  91.1  &   4.2x \\
Partial-only &  10  &  100\%  &  15.6  &  93.6  &   11.2x \\
Partial-only  &  50  &  5\%  &  15.9  &  88.9  &   3.0x \\
PACER  &  10  &  5\%  &  \textbf{16.1}  &  93.3  &  1.2x \\ 
PACER  &  10  &  33\%  &  15.9  &  94.6  & 3.8x \\
PACER  &  10  &  100\%  &  15.8  &  \textbf{96.3}  &  11.5x \\ 
\end{tabular}
}
\caption{\label{tab:pacer_perf}
Models (128-truncated) evaluated with various re-ranking schemes on the \textbf{validation} set. 
\textbf{Cost} is relative speed compared to no re-ranking at all.
}
\end{table}

\subsection{RPA Re-ranker Performance}

Table \ref{tab:full_char_classification_test} gives results for RPA-based re-ranking of generation models. 
Automated results show a slight bump in F1 on the LIGHT valid set, and indeed a bump in RPA. 
Including the intra-generation re-ranking with PACER yields an even higher RPA score. Table \ref{tab:pacer_perf} contains the results of varying the candidate tokens re-ranked per intra-generation step (\#Toks) and number of partial re-ranking steps (Freq), both in terms of generation metrics/RPA and relative computational cost compared to re-ranking. Increasing \# of toks or increasing the frequency can lead to improved F1 and RPA, but with significant latency increase for too high values (e.g. over 11x when applying re-ranking for every partial step using the top 10 tokens each time). Applying both partial and final complete ranking helps performance.

We note that for  re-ranker models the same model is both re-ranking outputs, and used to measure RPA afterwards, making that metric biased. Hence, human evaluations are required for this model, which will be detailed in Section
\ref{sec:human_eval}, which will indicate that re-ranking does in fact help.

\subsection{Unlikelihood}

Results of unlikelihood (UL) training are also given in Table \ref{tab:full_char_classification_test}.
We apply UL loss to the 128-truncation model in two different ways: (1) \textbf{Top-1}: apply the loss on the token that yields the most incorrect partial sequence RPA classification; (2) \textbf{All}: apply the loss to all tokens that yield an incorrect RPA classification on partial sequences. 
%
The RPA UL methods suffer compared to the baselines in terms of PPL and F1, yet they retain similar RPA metrics. We hypothesize that while the UL loss can adjust the model to refrain from generating out-of-character responses, there are still far too many other tokens that may yield similar outcomes that are not penalized. Table \ref{tab:unlikelihood_auto_valid} in Appendix \ref{sec:appendix_ul} includes similar results with the 1024-truncation model.

\subsection{Multi-Objective Training}

Multi-objective training results are in Table \ref{tab:multiobjective_vanilla_valid}, where the base model is a 1024-truncation model. We measure generation metrics in terms of RPA (with PPL and F1 in Table \ref{tab:multiobjective_vanilla_valid_appendix} in Appendix \ref{sec:appendix_mo}), and classification metrics in terms of Hits@1/427 as before. The model is able to predict the appropriate character using either the decoder outputs or the encoder+decoder outputs. 
For each case, $n_{MO}=2$ yielded better results than $n_{MO}=0$.
Interestingly, despite the relatively strong performance of the model in classifying the right character (87.42 hits@1 for the best model), this \textbf{does not} translate to substantial RPA improvements over the baseline. 

\begin{table}
\scriptsize
\centering
\resizebox{\linewidth}{!}{
\begin{tabular}{lll|rr}
\hline
\textbf{Input} & \textbf{{\tiny{$n_{MO}$}}} & \textbf{Stage} & RPA & Hits@1\\
\hline
Human & N/A & - & 92.8 & - \\
None & 0 & 0 & 88.4 & - \\
\hline
\multicolumn{5}{l}{\textbf{\textit{Multi-Objective}}} \\
Dec. only & 2  & 1 & 88.4 & 39.3 \\ 
Dec. only & 2 & 2 &  87.7 & 87.4 \\ 
Enc+Dec & 2 & 1 &  88.4 & 70.9 \\ 
Enc+Dec & 2 & 2 &  88.8 & 71.6 \\ 
\hline
\multicolumn{5}{l}{\textbf{\textit{Multi-Objective + Automated Expanded Attention}}} \\
Dec. Only & 0   & 1 & \textbf{89.1} & 86.4 \\
Dec. Only & 0  & 2 & \textbf{89.1} & \textbf{89.1} \\  
Enc+Dec & 2 & 1 &  88.4 & 83.3 \\
Enc+Dec & 2 & 2 & \textbf{89.1} & 88.5 \\
\end{tabular}
}
\caption{\label{tab:multiobjective_vanilla_valid}
Models trained with varying multi-objective setups, evaluated on the \textbf{valid} set. Models are initialized from a (1024-truncation) model fine-tuned on LIGHT.
}
\end{table}

\subsection{Expanded Attention}

\paragraph{Profile Grounding} Expanding the decoder attention yields significant gains across all automated metrics, as seen in Table \ref{tab:extra_attention_manual_valid} for a 1024-truncate model (and in Table \ref{tab:extra_attention_manual_valid_appendix} in Appendix \ref{sec:appendix_expanded_attention} for a 128-truncate model). As a baseline we explore simply re-attending to the full context again; this indeed improves metrics across the board for the short-context model, but the long-context model actually suffers. However, both models improve substantially over the baseline when including the full LIGHT context without the dialogue history, and attention over sub-components of the LIGHT context still yields strong improvements.

To see how much this expanded attention matters, we explored varying the number of rounds $r \in \{1, 2, 3\}$ of expanded attention, i.e., how many times the model attends to this additional context. In Table \ref{tab:extra_attention_manual_valid}, we also see that a second expanded attention round yields even better results, but performance drops off after applying a third round. 


\begin{table}
\small
\centering
\begin{tabular}{lr|rrr}
\hline
\textbf{Expanded Attention} & $r$ & \multicolumn{3}{c}{1024-Truncate Model}\\
& & PPL & F1 & RPA \\
\hline
Human & 0 & - & - & 92.80 \\
None  & 0 &  12.35 & 15.85 & 88.42 \\ 
\hline
ABCD + Dialogue Hist. & 1 &  12.47 & 15.82 & 88.34 \\
ABCD & 1 & 12.18 & \textbf{16.01} & 91.82 \\
ABCD & 2 &  \textbf{12.17}  &  15.95  &  \textbf{92.60}  \\
ABCD & 3 &   12.19  &  15.99  &  91.73  \\
\hline
ABC & 1 &  12.22 & 15.94 & 91.83\\
ABC & 2 &   12.24  &  15.99  &  92.24  \\
ABC & 3 &   12.25  &  15.93  &  92.25 \\
\hline
AB & 1 & 12.27  &  15.87  &  90.97  \\ 
A & 1 & 12.30  &  15.80  &  89.13  \\ 
B & 1 & 12.34  &  15.76  &  89.46 \\
\end{tabular}
\caption{\label{tab:extra_attention_manual_valid}
Models trained with expanded attention (profile grounding), evaluated on the \textbf{valid} set. Expanded attention input:
A = Self Persona, B = Self Name, C = Partner Name, D = Setting Description. We also vary the number of rounds $r$ of expanded attention. 
}
\end{table}

\paragraph{Automated Grounding}
We show results for the automated grounding of expanded attention in Table \ref{tab:extra_attention_automatic_valid}. Attempting to use the decoder attention weights to select expanded attention context yields no additional benefits, which is not surprising: if the model could identify the pertinent components of the input beforehand, it would not require a re-attention. The trainable mask does not yield any benefits either. However, using the RPA classifier attention weights to inform the model which tokens to re-attend to yields improved performance across all three metrics compared to the baseline,  and PPL is nearly the same as profile grounding (12.19 vs. 12.18), while RPA trails slightly behind (91.11 vs. 91.79). We also include the usage of the \textit{bottom-$k$} tokens from the classifier weights to emphasize that there is indeed signal from the top-$k$, as using the bottom tokens does not help.

\begin{table}
\centering
\resizebox{\linewidth}{!}{
\begin{tabular}{l|rrr}
\hline
\textbf{Exp. Attn. Grounding} &  PPL & F1 & RPA\\
\hline
Human & - & - & 92.80 \\
None  & 12.35 & 15.85 & 88.42 \\ 
\hline
Profile Ground Best (2 rounds) &  \textbf{12.17}  &  15.95  &  \textbf{92.60}  \\
Profile Ground Best (1 round) &  12.18  &  \textbf{16.08}  &  91.79  \\
Profile Ground Random & 12.43 & 15.74 & 87.62 \\
\hline
Decoder Attn. & 12.39 & 15.40 & 87.59 \\
Trainable Mask & 12.40 & 15.75 & 88.43 \\
Classifier Attn. (top-$k$) & 12.19 & 15.90 & 91.11 \\
Classifier Attn. (bottom-$k$) & 12.31 & 15.89 & 88.71 \\
\end{tabular}
}
\caption{\label{tab:extra_attention_automatic_valid}
Models trained with expanded attention (automated grounding), evaluated on the \textbf{valid} set. We vary the method for selecting the extra context to re-attend to. All models are long-truncation (1024).
}

\end{table}

\paragraph{Automated Grounding + Multi-Objective}  Table \ref{tab:multiobjective_vanilla_valid} shows that combining automated grounding with the multi-objective task yields higher hits@1 compared to not using the trainable mask, especially in the first stage of multi-objective training. However, 
RPA scores are only fractionally better than the baseline.
 Appendix \ref{sec:appendix_mo}  includes results across more settings (see Table \ref{tab:multiobjective_vanilla_valid_appendix} and Table \ref{tab:multiobjective_expanded_decoder_valid}).

\paragraph{Expanded Attention + RPA Re-ranking}

The expanded attention and RPA re-ranker methods  can also both be applied to obtain effective models. Results are in Table \ref{tab:full_char_classification_test}; indeed, the combination yields the highest F1 and RPA scores. 

\subsection{Human Evaluations}\label{sec:human_eval}

We performed human evaluation on our models. For each model we collected 100 human-model conversations, set up similarly to  the original LIGHT dataset conversations. During the conversation, crowdworkers were asked to annotate if the model's response: 1) contained a contradiction; 2) displayed a mistaken identity; 3) indicated a mistaken location; 4) was off-topic; 5) was repetitive; or 6) was all good. At the end of the conversation, the crowdworkers were asked to rate their partner's engagingness on a scale from 1-5. 

Results for mistaken identity, engagingness, and all-good responses are in Table \ref{tab:full_char_classification_test}, with full results (and comparison with a retrieval baseline) in  Table \ref{tab:human_evals} in the Appendix. The baseline model displays mistaken identity 6.45\% of the time, and has an average engagingness score of 4.04. Longer context increases engagingness to 4.16 but also increases mistaken identity. Unlikelihood and multi-objective training similarly increase mistaken identity. The successful methods, then, are the beam re-ranking methods and the expanded attention models. The long-context beam re-ranker decreases mistaken identity to 4.81\%, while the profile expanded attention model improves to 4\%, and has the best engagingness of 4.34. Combining RPA Re-ranking with expanded attention yields the lowest mistaken identity (2.38\%), while adding PACER leads to the highest all-good percentage (85.7\%). Correlations between automatic metrics and human evaluations are measured in Appendix \ref{sec:human_eval_correlation}, where we find that RPA and mistaken identity are indeed strongly correlated.

\section{Qualitative Analysis}

\subsection{Re-rankers \& Generation Settings}
\label{sec:reranker_analysis}



We further explored three decoding settings: standard beam-search, delayed beam search \citep{massarelli2019decoding} and nucleus sampling \citep{holtzman2019curious}, both in a re-ranking setting and not. When considering performance on automated metrics (provided in Table \ref{tab:gen_settings_valid} in the Appendix), we see that generation settings other than beam search, when using a re-ranker, yield lower F1 scores but higher RPA scores, as the RPA re-ranker has more diversity of candidate responses from which to choose; however, these methods perform worse in human evaluations, with nucleus sampling re-ranking yielding far more problems and far lower engagingness ratings. Qualitative analysis of outputs on the test set in Appendix \ref{sec:reranker_analysis_appendix}.

\subsection{Left-to-Right Dynamic Classification}

We find that the left-to-right RPA classifiers are correctly sensitive to per-token perturbations in the input, and can accurately predict the speaker at the token level. In Table \ref{tab:ltr_dynamic_class}, we give an example where the classifier changes its character prediction, depending on the candidate utterance.

\begin{table*}[t]
\begin{center}
\resizebox{\linewidth}{!}{
\begin{tabular}{|lll|lll|}
\hline
\multicolumn{6}{|p{1.0\linewidth}|}{\textbf{Setting:} Turquoise Shore, Shore} \\
\multicolumn{6}{|p{1.0\linewidth}|}{A beautiful turquoise color water by the shore. It is filled with many gems and gold.} \\
\multicolumn{6}{|p{1.25\linewidth}|}{\textbf{Character 1:} \textbf{Sea Witch}. I am a sea witch.  I pray on young sailors who hope to find adventure and treasures on the open sea.  I lure them in with magic spells and promise of riches.} \\
\multicolumn{6}{|p{1.25\linewidth}|}{\textbf{Character 2:} \textbf{Mermaid}. I am one of the most beautiful mermaids to live in the sea. I like to watch the other sea creatures swim by me, including dolphins, who are my favorite creatures because they are so friendly. I fear the people who live on land because they hunt my kind} \\
\hline
\multicolumn{3}{|l|}{\textbf{Classified Utterance: } Hey there Mermaid! Long time, no see.} & \multicolumn{3}{|l|}{\textbf{Classified Utterance: } Hey there Sea Witch! Long time, no see.} \\
\multicolumn{3}{|l|}{\textbf{Correct Speaker: } Sea Witch} & \multicolumn{3}{|l|}{\textbf{Correct Speaker:} Mermaid} \\
\hline
\textbf{Word} & \textbf{Predicted Speaker} & \textbf{Confidence} & \textbf{Word} & \textbf{Predicted Speaker} & \textbf{Confidence}\\
Hey & sea witch &  0.5156 & Hey & sea witch & 0.5156 \\
there & sea witch &  0.5467 & there & sea witch &  0.5467 \\
Mermaid! & sea witch &  0.9978 & sea & mermaid & 0.9968 \\
& & & witch & mermaid & 1.000 \\
Long & sea witch &  0.9981 & Long & mermaid & 1.000 \\
time, & sea witch &  0.9979 & Time & mermaid & 1.000 \\
no & sea witch &  0.9982 & no & mermaid & 1.000 \\
see. & sea witch &  0.9985 & see. & mermaid & 1.000 \\
\hline
\end{tabular}
}
\end{center}
\caption{Left-to-right dynamic classification examples. A candidate utterance is shown, along with the classifier's predictions at each partial decoded sequence. \textbf{Left}: The true next utterance in the dialogue, with the RPA classifier's predictions and confidence token by token. \textbf{Right}: A perturbed utterance. If we switch the name being addressed, the model switches its predictions immediately.}
\label{tab:ltr_dynamic_class}
\end{table*}

\subsection{Classifier Failure Modes}
\label{sec:rpa_failure_modes}

We note that the human dialogue data is classified as being ``in character'' only 92.8\% of the time on the validation set by the LTR RPA classifier. We examine the scenarios in which the classifier is incorrect, with example input/output pairs in Table \ref{tab:human_incorrect} in the Appendix. First, there are instances where either character could have said the output response (row 1). Second, there are instances where there are not enough clues in the context to provide an estimation of who said the response, for example at the beginning of the conversation (row 2). And, there are still some small amount of instances that the classifier simply fails (row 3).

\subsection{Per-Turn Character Accuracy Analysis}

\begin{figure}[h!]
    \centering
    \includegraphics[width=\linewidth]{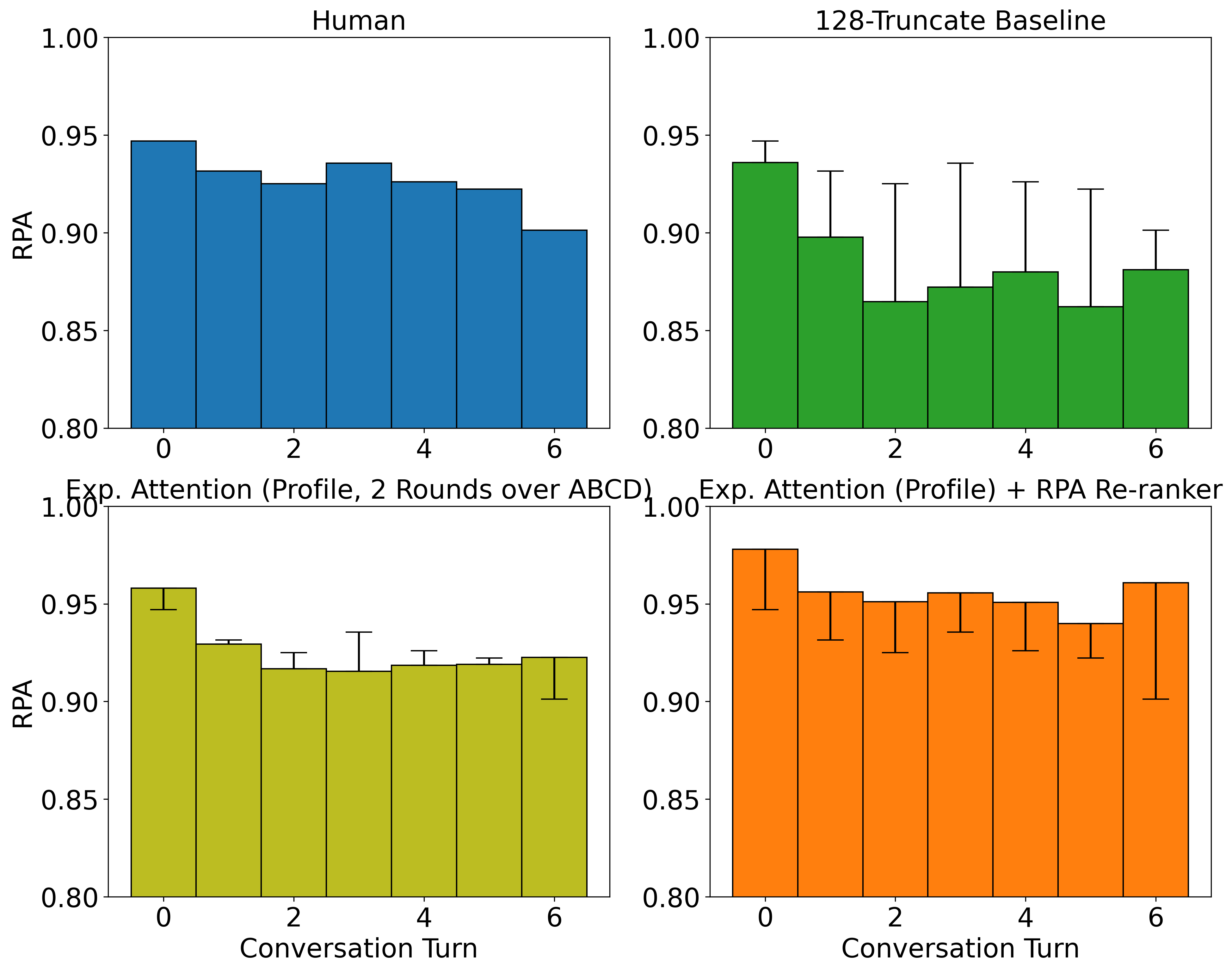}
    \caption{Per-turn RPA classifications, for a variety of models. Error bars show the difference between the model's RPA value and the human's RPA value.}
    \label{fig:per_turn_analysis_main}
\end{figure}

We consider the RPA of various models when evaluated across the turns of conversation. Intuitively, baseline models would suffer as the conversation goes on for a variety of reasons (character roles are truncated out of context, more input yields noisier outputs, etc.). In Figure \ref{fig:per_turn_analysis_main}, we display the per-turn results for a few representative models, with Figure \ref{fig:per_turn_analysis_appendix} in Appendix  \ref{sec:per_turn_analysis_appendix} enumerating over a more comprehensive set. The \textbf{human} outputs are most often correct on the first turn, with gradual RPA decay throughout the conversation. The \textbf{128-truncate} baseline, as expected, suffers a dramatic performance drop after the first couple of turns. Meanwhile, with the \textbf{profile expanded attention}, we see near-human performance, with better RPA in later turns. Including RPA re-ranking improves dramatically over all turns.

\subsection{Expanded Attention Visualization}

To gain some insight into what is happening with the expanded attention, we mapped out the attention between context and response tokens for both a baseline model with no expanded attention, and a model with profile expanded attention. Figure \ref{fig:no_extra_attention} and Figure \ref{fig:manual_attention} in the Appendix display the heat maps for an example context and response. Information about the heat map construction and the specific example used are in Appendix \ref{sec:heatmap_appendix}.

The baseline model spreads its attention out across both the LIGHT context and the dialogue history, with the majority of the attention looking at overlapping words in the context and the response and almost no attention on the character names. The expanded attention model concentrates the first attention on the recent dialogue history heavily in the first level of attention, and then concentrates on pertinent words in the context related to the character information (i.e., the character names) in the second round of attention. 

\section{Discussion \& Conclusion}

In this work we explored the  problem of maintaining one's character in open dialogue, and
showed that state-art-of-the-art models have a fundamental weakness in this regard.
We provided a clear framing of the problem and showed one can build automatic metrics  (RPA) that evaluate models using a classifier. We then explored a variety of methods throughout this paper; we offer a brief discussion of the most crucial takeaways here. 

\paragraph{RPA Classifiers} The RPA classifiers are effective in both assigning identities to unknown speakers, and measuring the role-playing effectiveness of candidate models. Those robust to partial sequences are even more effective as they can be utilized in tools requiring such partial classification, including unlikelihood training and re-ranking methods.


\paragraph{Alternative Training Methods} The alternative training methods did not yield improved results, as hoped. \textbf{Unlikelihood} training does not improve RPA. 
 \textbf{Multi-objective} training shows that we can predict the character from the model's internal representations, but  does not seem to improve downstream generation results. It remains an open problem how to integrate improvements into training.

\paragraph{Expanded Attention} The straightforward application of additional re-attention to relevant contextual inputs yields substantial gains across the board, both in automated metrics and human evaluations. Intuitively this makes sense; we are telling the model that it cannot decode its response without first being reminded of its character. Automated attempts to choose this context using the RPA classifier yielded improvements in RPA compared to the baseline, re-affirming that the RPA classifiers pay attention to pertinent tokens in the input; in human evaluations the method displayed lower frequency of mistaken identity as well, though not as low as profile grounding.

\paragraph{Re-rankers} The re-rankers help improve RPA measurements and indeed show stronger performance in resolving mistaken identity issues in human evaluations. As these methods are model-agnostic, we can combine with expanded attention, which yields the highest RPA metrics and lowest frequency of mistaken identity issues, demonstrating its effectiveness. PACER performs well and may be suitable for other tasks beyond the focus of this paper.

\paragraph{Comparison with Human Performance}
Our best methods still lag behind human (crowdworker) performance in several regards, e.g. 1.34\% vs. 2.23\% in terms of mistaken identity per turn, or 5\% vs. 14.7\% per conversation. Therefore considerable progress still has to be made on this challenging problem.

\bibliography{custom}
\bibliographystyle{acl_natbib}

\appendix

\section{Training Details}
All models are trained with the ParlAI framework \citep{miller2017parlai}.

\label{sec:training_details}
\paragraph{Base Models} RPA classifier Poly-encoders are initialized with the 622M parameter models from \citet{roller2020recipes}; we also use this architecture for dialogue response (retrieval) models which we also evaluate (see Table \ref{tab:human_evals}). All generative models are initialized with BlenderBot, also from \citet{roller2020recipes}, a 2.7B parameter transformer encoder/decoder model. Each model was pre-trained on 1.5B training examples from pushshift.io Reddit \citep{baumgartner2020pushshift}, with BlenderBot additionally fine-tuned on the BST tasks (see \citet{roller2020recipes} for more details), before training on LIGHT.

\paragraph{RPA Classifiers} The RPA classifier models are trained with a cross-entropy loss over the correct label, with 99 random negatives chosen from the training set; we ensured that each character in conversation showed up in the set of candidate labels. The models were trained with a batch size of 16 on 4 32GB GPUs, with early stopping on the validation set according to valid accuracy. We used the Adam optimizer \citep{kingma2014adam} with weight decay \citep{loshchilov2018decoupled}, sweeping over learning rates $\{1e-5, 5e-6\}$.
\paragraph{Generative Models} All variants of generative models were trained using 8 32GB GPUs, with early stopping on perplexity on the validation set. We used the Adam optimizer, sweeping over learning rates $\{1e-5, 7e-6\}$, training with a batch size of 128 for the short-truncation models, and 32 for the long-truncation models. For the multi-objective models, we used the same loss (and negative-sampling) setup as the RPA classifiers for the character accuracy objective. During inference, unless otherwise specified, we generated using beam-search with beam size of 10, enforcing a minimum length of 20, and with tri-gram blocking with respect to both the context and the current generation.

\section{RPA Classifier Training}
\begin{table}
\small
\centering
\resizebox{\linewidth}{!}{
\begin{tabular}{l|r|r|r}
\hline
\textbf{Dataset} & \multicolumn{1}{c}{Train} &  \multicolumn{1}{c}{Valid} &  \multicolumn{1}{c}{Test}\\
\hline
LIGHT {\tiny \citep{urbanek2019learning}} & 111k & 6k & 13k \\
\hline
RPA, 0-Utterance & 212k & 12k & 26k \\
RPA, 4-Utterance & 748k & 45k & 90k \\
RPA, All-Utterance & 34k & 2k & 4k \\
\hline
\hline
RPA LTR, 0-Utterance & 3.3M & 205k & 414k\\
RPA LTR, 4-Utterance & 12M & 747k & 1.5M \\
RPA LTR, All-Utterance & 516k & 31k & 64k \\
\hline
\end{tabular}
}
\caption{\label{tab:dataset_stats}
Number of training, valid, and test examples for the LIGHT dataset, as well as the RPA training splits (both normal and LTR).
}
\end{table}

\label{sec:rpa_train_data_appendix}

We build the training data for the RPA classifiers from the LIGHT dataset. The input is a concatenation of (1) the LIGHT context (set of characters, setting, etc.); (2) a fixed number of previous utterances in the conversation; and (3) a candidate utterance from \textit{any point later} in the conversation (a special 
token separates the candidate utterance from the prior context). We experiment with either 0, 4, or $N-2$ prior utterances (dubbed ``All'' in relevant tables), where $N$ is the total number of utterances ($N-2$ allows the last turn for each speaker to be a candidate utterance). The left-to-right (LTR) data split is built similarly, except each example $i$ becomes $w_i$ examples, where $w_i$ is the number of tokens in the candidate utterance for example $i$. 
Statistics of the training dataset are given in Table \ref{tab:dataset_stats}.


Suppose we choose $n$ as the number of prior utterances to include in the input, and let us denote $D=8538$ to represent all the dialogues in the LIGHT train split, and $U = 110877$ to represent all the utterances in those dialogues. For the RPA classification dataset, each dialogue is presented twice, once from each character's POV. When $n = N-1$, where $N$ is the length of a conversation, then we have roughly $2D$ training examples. When $n = 0$, we have roughly $2U$ training examples. 

For any value $0 < n < N-1$, we build out several examples from several slices of each conversation. Suppose we have dialogue $d_i$ with $N$ utterances $\{u_0, u_1, ..., u_N\}$. To build the training data from dialogue $d_i$, we select all continuous subsets of $n$ utterances within $d_i$, forming contexts $$c_i = \{u_{i},...,u_{i+n}\}\quad\forall\quad 0 \leq i \leq N-i$$ Then, we look at all $N - i$ utterances following utterance $u_{i+n}$, and use these as target utterances in the task. The goal of this is to build the model to be robust to dataset artifacts; without this modification, the model could trivially pick out the character just by looking at the number of alternating utterances. These measures force the model to fully understand the task and react accordingly.

\section{RPA Classifier Performance: Additional Results}
In Table \ref{tab:character_classifiers_valid_full}, we see how each RPA classifier performs on the various datasplits, varying the number of prior utterances used during training and evaluation. Each model performs best on the split on which it was trained (the highlighted numbers).

\begin{table}
\small
\centering
\begin{tabular}{l|rrr}
\hline
\textbf{\# Train-time} & \multicolumn{3}{c}{\textbf{Hits@1/427}}\\
Prior Utterances & \multicolumn{3}{c}{\# Eval Prior Utterances} \\
& 0 & 4 & All \\
\hline
\multicolumn{4}{c}{\textbf{Without LIGHT Context}} \\
\hline
0 & \win{10.35} & 18.58 & 17.71 \\
4 & 2.10 & \win{87.31} & 84.35\\
All & 7.02 & 81.26 & \win{85.70} \\
\hline
\multicolumn{4}{c}{\textbf{With LIGHT Context}} \\
\hline
0 & \win{77.64} & 66.20 & 58.61 \\
4 & 31.04 & \win{86.48} & 84.90 \\
All & 32.54 & 82.73 & \win{89.26} \\
\hline
\end{tabular}
\caption{\label{tab:character_classifiers_valid_full}
RPA classifier performance on the \textbf{validation} set, as measured by hits@1/427.
Highlighted numbers indicate models evaluated on the split on which they were trained.
}
\end{table}

\section{Unlikelihood: Additional Results}
\label{sec:appendix_ul}

In Table \ref{tab:unlikelihood_auto_valid}, we compare UL models across different truncation lengths; the same story applies to the 1024-truncation models. We additionally include a third method, \textbf{Random-3}, where we apply the loss randomly to 3 tokens that yield incorrect RPA classifications. This method performs about the same as the Top-1 method, but the RPA is lower, indicating that the Top-1 method at least is providing some signal.

\begin{table}
\small
\centering
\resizebox{\linewidth}{!}{
\begin{tabular}{l|rrr}
\hline
\textbf{Unlikelihood Method} & PPL & F1 & RPA\\
\hline
Human & 1 & 1 & 92.80 \\
None (128) &  12.54  &  15.80  &  88.54  \\ 
None (1024) &  \textbf{12.35} & \textbf{15.85} & 88.42  \\ 
\hline
\textbf{128-Truncation} \\
RPA UL: Top-1 Token  &  13  &  15.35  &  88.54  \\ 
RPA UL: All tokens  &  12.86  &  15.28  &  \textbf{88.86}  \\ 
RPA UL: Random-3  &  12.99  &  15.37  &  87.85  \\ 
\textbf{1024-Truncation} \\
RPA UL: Top-1 Token  &  12.49  &  15.66  &  88.12  \\ 
RPA UL: All tokens  &  12.57  &  15.83  &  88.06  \\ 
\hline
\end{tabular}
}
\caption{\label{tab:unlikelihood_auto_valid}
Models trained with unlikelihood loss, evaluated on the \textbf{valid} set. We vary the tokens to which we apply UL loss.
}
\end{table}

\section{Multi-Objective: Additional Results}
\label{sec:appendix_mo}

\subsection{Perplexity \& F1}
Table \ref{tab:multiobjective_vanilla_valid_appendix} displays full PPL and F1 scores corresponding to the models in Table \ref{tab:multiobjective_vanilla_valid}.

\begin{table}
\scriptsize
\centering
\resizebox{\linewidth}{!}{
\begin{tabular}{lll|rrrr}
\hline
\textbf{Input} & \textbf{{\tiny{$n_{MO}$}}} & \textbf{Stage} & PPL & F1 & RPA & Hits@1\\
\hline
Human & N/A & & - & - & 92.8 \\
None & 0 & 0 & \textbf{12.4} & 15.9 & 88.4 \\
\hline
\multicolumn{7}{l}{\textbf{\textit{Multi-Objective}}} \\
Dec. only & 2  & 1 &  \textbf{12.4}  &  15.9  &  88.4 & 39.3 \\ 
Dec. only & 2 & 2 &  12.8  &  \textbf{16.0}  &  87.7 & 87.4 \\ 
Enc+Dec & 2 & 1 &  \textbf{12.4}  &  15.9  &  88.4 & 70.9 \\ 
Enc+Dec & 2 & 2 &  12.5  &  15.8  &  88.8 & 71.6 \\ 
\hline
\multicolumn{7}{l}{\textbf{\textit{Multi-Objective + Automated Expanded Attention}}} \\
Dec. Only & 0   & 1 & 13.2 & 15.7 & \textbf{89.1} & 86.4 \\
Dec. Only & 0  & 2 &  12.9 & 15.9 & \textbf{89.1} & \textbf{89.1} \\  
Enc+Dec & 2 & 1 &  12.9 & 15.8 & 88.4 & 83.3 \\
Enc+Dec & 2 & 2 & 12.7 & 15.8 & \textbf{89.1} & 88.5 \\
\end{tabular}
}
\caption{\label{tab:multiobjective_vanilla_valid_appendix}
Models trained with varying multi-objective setups, evaluated on the \textbf{valid} set. Models are initialized from a (1024-truncation) model fine-tuned on LIGHT.
}
\end{table}

\subsection{Multi-Objective + Automated Grounding}
In Table \ref{tab:multiobjective_expanded_decoder_valid}, we see additionally how, when using either the encoder+decoder or just the decoder outputs, we do not require additional multi-objective layers (as we did in the non-automated-grounding case).

\begin{table}
\resizebox{\linewidth}{!}{
\begin{tabular}{lll|rrrr}
\hline
\textbf{Input} & $n_{MO}$ & \textbf{Stage} & PPL & F1 & RPA & Hits@1/427\\
\hline
Human & 0 & & 1 & 1 & 92.80 \\
None & 0 & 0 & \textbf{12.35} & 15.85 & 88.42 \\
\hline
Dec. Only & 0   & 1 & 13.22 & 15.66 & 89.08 & 86.37 \\
Dec. Only & 0  & 2 &  12.92 & \textbf{15.88} & 89.10 & 89.10 \\  
Enc+Dec & 0  & 1 &  13.24 & 15.55 & 88.83 & 85.78 \\
Enc+Dec & 0  & 2 &  13.44 & 15.61 & \textbf{89.29} & \textbf{89.22} \\
Enc+Dec & 2 & 1 &  12.94 & 15.80 & 88.39 & 83.25 \\
Enc+Dec & 2 & 2 & 12.69 & 15.77 & 89.05 & 88.49 \\
\end{tabular}
}
\caption{\label{tab:multiobjective_expanded_decoder_valid}
Models trained with varying multi-objective + automated grounding setups, evaluated on the \textbf{valid} set. The base model in all cases is initialized from a generation model fine-tuned on LIGHT.
}
\end{table}

\section{Expanded Attention: Additional Results}
\label{sec:appendix_expanded_attention}

We provide results for both the 128-truncate and 1024-truncate models with profile grounding in Table \ref{tab:extra_attention_manual_valid_appendix}. Trends remain the same for both models.

\begin{table}
\scriptsize
\centering
\begin{tabular}{lr|rrr|rrr}
\hline
\textbf{Exp.} & $r$ & \multicolumn{3}{c}{128-Truncate Model} & \multicolumn{3}{c}{1024-Truncate Model}\\
\textbf{Attn.} & & PPL & F1 & RPA & PPL & F1 & RPA \\
\hline
Human & 0 & - & - & 92.80 & 1 & 1 & 92.80 \\
None  & 0 & 12.59  &  15.80  &  88.28  & 12.35 & 15.85 & 88.42 \\ 
\hline
ABCD+ & 1 & \textbf{12.23} & 15.87 & 90.59 & 12.47 & 15.82 & 88.34 \\
ABCD & 1 & 12.25 & 15.97 & 90.94 & 12.18 & \textbf{16.01} & 91.82 \\
ABCD & 2 & \textbf{12.23}  &  15.89  &  90.83  &  \textbf{12.17}  &  15.95  &  \textbf{92.60}  \\
ABCD & 3 & 12.26  &  15.81  &  90.44  &  12.19  &  15.99  &  91.73  \\
\hline
ABC & 1 & 12.33 & 15.82 & 91.50 & 12.22 & 15.94 & 91.83\\
ABC & 2 & 12.31  &  16.03  &  \textbf{92.03}  &  12.24  &  15.99  &  92.24  \\
ABC & 3 & 12.33  &  15.90  &  91.59  &  12.25  &  15.93  &  92.25 \\
\hline
AB & 1 &  12.42  &  15.92  &  90.31  &  12.27  &  15.87  &  90.97  \\ 
A & 1 &  12.46  &  \textbf{16.05}  &  90.22  &  12.30  &  15.80  &  89.13  \\ 
B & 1 &  12.53  &  15.85  &  89.85  &  12.34  &  15.76  &  89.46 \\
\end{tabular}
\caption{\label{tab:extra_attention_manual_valid_appendix}
Models trained with expanded attention (profile grounding), evaluated on the \textbf{valid} set. Expanded attention input:
A = Self Persona, B = Self Name, C = Partner Name, D = Setting Description, + = dialogue history. We also vary the number of rounds $r$ of expanded attention. 
}
\end{table}

\begin{table*}
\centering
\scriptsize
\begin{tabular}{l|rrr}
\hline
 & Percentage &  & \\
Model: & Beam Baseline & Delayed Beam Baseline & Delayed Beam with Re-ranker\\
\hline
thinks it is someone else/partner & 0\% & 13.04\% & 19.23\%\\
Thinks partner's character is its character (i.e., thinks it is talking to itself) & 57.69\% & 56.52\% & 11.54\%\\
emulates partner's characteristic & 0\% & 4.35\% & 0\%\\
incorrectly identifies partner & 19.23\% & 17.39\% & 30.77\%\\
talks about its character in the 3rd person & 0\% & 4.35\% & 0\%\\
emulates irrelevant characteristic & 3.85\% & 0\% & 7.69\%\\
combines self and partner persona & 7.69\% & 0\% & 9.62\%\\
incorrectly identifies 3rd party character & 0\% & 0\% & 1.92\%\\
claims it does not know who it is & 0\% & 0\% & 1.92\%\\
noise & 11.54\% & 4.35\% & 17.31\% \\
\hline
\end{tabular}
\caption{\label{tab:turn_annotation_manual_analysis}
Turn annotation analysis of RPA Re-rankers.
}
\end{table*}

\section{Full Valid Results}
Table \ref{tab:full_char_classification_valid} includes results on the LIGHT validation set for models in Table \ref{tab:full_char_classification_test}.

\begin{table*}
\small
\centering
\begin{tabular}{l|rrr}
\hline
\textbf{Model} & PPL & F1 & RPA\\
\hline
Human & 1 & 1 & 92.80 \\
\hline
\multicolumn{4}{l}{\textbf{\textit{Baselines}}} \\
128-Truncate Vanilla Baseline & 12.54 & 15.80 & 88.54 \\
1024-Truncate Vanilla Baseline & 12.35 & 15.85 & 88.42 \\
\hline
\multicolumn{4}{l}{\textbf{\textit{Re-rankers}}} \\
128-Truncate Baseline + Re-Ranker & - & 16.14 & 92.99 \\
128-Truncate Baseline + PACER & - & 16.13 & 93.31\\
\hline
RPA UL (Top-1 Token) & 13.00 & 15.35 & 88.54 \\
RPA UL (All Tokens) & 12.86 & 15.28 & 88.86 \\
Multi-Objective (Vanilla, Dec. Only)  & 12.78 & 16.00 & 87.71 \\
\hline
\multicolumn{4}{l}{\textbf{\textit{Expanded Attention Methods}}}\\
Profile Grounding (128, 2 Rounds over ABC) &12.31 &  16.03 & 92.03 \\
Profile Grounding (1024, 2 Rounds over ABCD) & 12.17 & 15.95 & 92.60\\
Automated Grounding (1024, Classifier Attn.) & 12.19 & 15.90 & 91.11\\
Automated Grounding + MO (1024 Dec. Only) & 12.92 & 15.88 & 89.10\\
\hline
\multicolumn{4}{l}{\textbf{\textit{Expanded Attention + Re-ranker Methods}}}\\
Profile (128) + RPA Re-ranker & - & 16.21 & 95.62\\
Profile (128) + PACER & - & 16.18 & 95.82 \\
\hline
\end{tabular}
\caption{\label{tab:full_char_classification_valid}
Validation statistics for various models considered throughout the paper.
}
\end{table*}

\section{Retrieval Re-rankers}

We evaluated a Poly-encoder baseline model with an RPA re-ranker as well. The Poly-encoder scores utterances from the full training set as candidates, and the candidates for re-ranking are the top-$k$ ranked utterances; results are in Table \ref{tab:retrieval_reranker_valid}. Retrieval models benefit dramatically from the re-ranking, improving to almost 99\% RPA as measured by the LTR classifier. As the candidate responses for retrieval models come from the set of all training utterances, and due to overlap between the set of characters appearing in the train and valid sets, we can examine how often the model output was originally spoken by its partner's character; this can be seen as a proxy for mistaken identity. We find that the re-ranker reduces the amount of time that the model returns a message its partner said, indicating some viable and promising results.

\begin{table}
\small
\centering
\begin{tabular}{l|r|r}
\hline
\multicolumn{1}{c}{\textbf{Metric}} & \multicolumn{1}{c}{Baseline} & \multicolumn{1}{c}{Re-ranker}\\
\hline
RPA (normal) & 85.47 & 94.29 \\
RPA (LTR) & 86.31 & 99.76 \\
\% Partner Said Response & 3.20 & 2.02 \\
\hline
\end{tabular}
\caption{\label{tab:retrieval_reranker_valid}
Retrieval models with character output re-rankers; performance on the \textbf{validation} set.
}
\end{table}

\section{Full Human Evaluation Results}
In Table \ref{tab:human_evals}, we display the full results of human evaluations across all dimensions. We note that the Poly-encoder model is best at not mistaking location or being repetitive, but this is expected given its retrieving over human-written utterances.

\begin{table*}[h!]
\centering
\resizebox{\linewidth}{!}{
\begin{tabular}{l|rrrrrr|rr|r}
\hline
\textbf{Model} & Contradiction  & Mistaken & Mistaken & Off-Topic & Repetitive & All-Good & Clean & Mistaken & Avg.  \\
& & Identity & Location  & & & & Convo & Identity & Engagingness\\
& &          &           & & & &       & In Convo & \\
\hline
Human & - & 1.34\% & - & - & - & - & - & 5\% & - \\
\hline
\textit{\textbf{Baselines}} \\
Poly-Encoder & 5.50\% & 6.14\% & \textbf{0.77\%} & 12.02\% & \textbf{1.92\%} & 75.45\% & 16.33\% & 34.69\% & 3.42  \\
128-Truncate Vanilla Baseline & 8.26\% & 	6.45\% & 	2.71\% &	4.26\% &	4.00\% &	76.00\%	& 26.80\% & 35.05\% & 4.04 \\
1024-Truncate Vanilla Baseline & 7.48\% & 7.35\% & 2.66\% & 6.21\% & 4.31\% & 75.03\% & 22.22\% & 38.38\% & 4.16  \\
\hline
\textbf{\textit{Re-rankers}} \\
128-Truncate Baseline + RPA Re-Ranker (Beam) & 4.83\% & 5.56\% & 	3.62\%	& 4.35\%	& 3.26\% & 	80.31\% &	20.19\% & 34.65\% & 4.14 \\
128-Truncate Baseline + RPA Re-Ranker (Nucleus) & 9.07\% & 	8.68\% & 	2.33\% & 	5.31\% & 	3.89\% & 	73.70\% & 	31.96\% & 37.11\% & 	3.83 \\
1024-Truncate Baseline + RPA Re-Ranker (Beam) & 5.55\% & 4.81\% & 1.60\% & 3.45\% & 2.71\% & 82.98\% & 33.33\% & 24.45\% & 4.14 \\
128-Truncate Baseline + PACER & 8.28\% & 	4.27\% & 	4.89\% & 	3.14\% & 	3.14\% & 	73.90\% & 	21.78\% & 33.66\% & 	3.96 \\
1024-Truncate Baseline + PACER & 7.63\%  &  7.13\%  &  2.38\%  &  3.63\%  &  3.75\%  &  79.25\%  &  28.00\% & 36.00\%  &  4.18 \\ 
\hline
\textbf{\textit{Modified Training Objectives}} \\
RPA Unlikelihood (Top-1 Token) & 8.70\% & 7.13\% & 3.38\% & 7.25\% & 3.74\% & 72.83\% & 14.42\% & 39.40\% & 3.87  \\
RPA Unlikelihood (All Tokens) & 11.64\% & 	10.51\% & 	3.13\% & 	4.88\% & 	5.38\% & 	67.71\% & 	19.00\% & 43.00\% & 	3.87 \\
Multi-Objective (Vanilla, Dec-Only) & 8.13\% & 10.00\% & 	1.88\% & 5.63\%	 & 2.63\% & 74.75\%	& 18.00\% & 49.00\% & 	4.21\\
\hline
\textbf{\textit{Expanded Attention Methods}} \\
Profile Grounding (128, 2 Rounds over ABC) & 5.32\% & 	4.82\% & 	3.21\% & 	4.45\% & 	2.84\% & 	81.58\% & 	27.45\% & 28.43\% &  	4.18\\
Profile Grounding (1024, 2 Rounds over ABCD) & \textbf{4.13\%} & 4.00\% & 3.38\% & \textbf{3.13\%} & 3.25\% & 83.75\% & 36.63\% & 23.76\% & 4.34  \\
Automated Grounding (Classifier Attn.) & 10.17\% & 5.51\% & 2.57\% & 6.13\% & 2.33\% & 75.98\% & 24.27\% & 29.13\% & 4.04 \\
Automated Grounding + MO (Dec. Only) & 8.23\% & 	4.43\% & 	2.03\% & 	3.80\% & 	5.19\% & 	78.61\% & 	38.00\%  & 23.00\% & 	4.12\\
\hline
\textbf{\textit{Expanded Attention + Re-Ranker Methods}} \\
Profile Grounding (128) + RPA Re-Ranker & 5.33\% & 	\textbf{2.23\%} & 	1.61\% & 	4.22\% & 	2.98\% & 	84.37\% & 	36.27\% & \textbf{14.71\%} & 	4.25 \\
Profile Grounding (1024) + RPA Re-Ranker & 6.00\%  &  3.60\%  &  1.20\%  &  \textbf{0.42\%}  &  \textbf{0.30\%}  &  85.25\% &  40.00\% & 21.90\%  &  \textbf{4.35} \\
Profile Grounding (128) + PACER  & 5.43\% & 	4.07\% & 	2.84\% & 	2.47\% & 	1.23\% & 	\textbf{85.70\%} & 	\textbf{41.18\%} & 24.51\% & 	4.32 \\
Profile Grounding (1024) + PACER  & 6.21\% & 	4.38\% & 	1.10\% & 	3.65\% & 	2.56\% & 	83.56\% & 	40.78\% & 22.33\% & 	4.13 \\
\hline
\end{tabular}
}
\caption{\label{tab:human_evals}
Human evaluations. Annotators chatting with models were asked to annotate whether model utterances contained any of the problem attributes listed, with ``All-Good'' indicating that there were no issues. ``Clean Convo'' is the percentage of conversations without any issues.
}
\end{table*}

\section{Generation Settings}

\subsection{Test Output Analysis}
\label{sec:reranker_analysis_appendix} 
We provide qualitative analysis of the various generation methods below.

\paragraph{No Re-Ranking} When examining the baseline with no re-ranking, we found that nucleus sampling can help when beam search does not work; however, both can go out of character the farther one goes in conversation. 
\paragraph{Beam Search Re-Rankers} The beam outputs in standard beam search are at times too similar, in which case re-ranking does next to nothing, unless a viable response is available.
\paragraph{Nucleus Sampling Re-Rankers} Using nucleus setting in a re-ranking setup yields more diverse choices to choose from; however, sometimes the model simply does not address *any* character within the conversation.
\paragraph{Delayed Beam Search Re-Rankers} This strikes a nice balance between sensible outputs from beam search and diversity from nucleus sampling.
\paragraph{Mixed-Decoding Re-ranker} Using mixed decoding (re-ranking several decoding schemes) can work quite well, as it is a nice blend of different generation methods.

\subsubsection{Turn Annotation Analysis}
\label{sec:reranker_turn_analysis}

Qualitative analysis of the turn annotation results are in Table \ref{tab:turn_annotation_manual_analysis}. We generally found that beam search fails the vast majority of the time when the model thinks that it is talking to \textit{itself}; i.e., it confuses its partner for its own character. The re-rankers can help shift the hallucination away from this regime.

\subsection{Automated Metrics}
We experiment with various generation settings, with or without re-rankers; results are in Table \ref{tab:gen_settings_valid}. For the baseline and re-ranker models, beam search yields the highest F1 scores; RPA can be improved with the other inference methods when combined with a re-ranker. We believe this may be due to the higher diversity of candidate responses generated from those methods.

\begin{table*}
\small
\centering
\begin{tabular}{l|rr|rr}
\hline
 & \multicolumn{2}{c|}{\textbf{Normal}} & \multicolumn{2}{c}{\textbf{Re-ranking}}\\
\textbf{Generation Setting} &  F1 & RPA & F1 & RPA\\
\hline
Human & 1 & 1 & 92.80 \\
\hline
\multicolumn{3}{l|}{\textit{\textbf{128-Truncation Model}}} \\
Beam Search &   15.80  &  88.54 & \textbf{16.14} & 92.99 \\ 
Delayed Beam Search &  15.46 & 88.74 & 15.48 & 93.18 \\ 
Nucleus Sampling & 15.70 & \textbf{89.25} & 15.44 & 97.12 \\ 
Top-K Sampling & 14.47 & 88.16 & 14.14 & 97.01 \\ 
\hline
\multicolumn{3}{l|}{\textit{\textbf{1024-Truncation Model}}} \\
Beam Search  & \textbf{15.85} & 88.42 & 16.08 & 92.92  \\ 
Delayed Beam Search & 15.03 & 88.00 & 15.39 & 92.89 \\ 
Nucleus Sampling & 15.42 & 88.22 & 15.25 & \textbf{97.24} \\ 
Top-K Sampling & 14.45 & 86.91 & 14.06 & 97.15 \\ 
\hline
\end{tabular}
\caption{\label{tab:gen_settings_valid}
Performance on the LIGHT \textbf{valid} set for the baseline models with different generation settings, with or without re-rankers. All settings use tri-gram blocking with respect to the context and current generation, and have a minimum length of 20. We set $topp = 0.3$ for Nucleus sampling, $topk = 50$ for Top-K sampling, and use a beam-delay of 10 with $topk=10$ for delayed beam search.
}
\end{table*}

\begin{table*}[t]
\begin{center}
\scriptsize
\begin{tabular}{|p{0.9\linewidth}|}
\hline
\textbf{Context:} \_setting\_name Turquoise Shore, Shore \\
\_setting\_desc A beautiful turquoise color water by the shore. It is filled with many gems and gold.\\
\_partner\_name sea witch\\
\_self\_name mermaid\\
\_self\_persona I am one of the most beautiful mermaids to live in the sea. I like to watch the other sea creatures swim by me, including dolphins, who are my favorite creatures because they are so friendly. I fear the people who live on land because they hunt my kind.\\
\textbf{Dialogue History: }\\
Hey there Mermaid! Long time, no see.\\
Long time indeed! How have you been keeping?\\
Pretty good, You know how it goes. Just trying to find some unwitting victims. What are you doing in the Turquoise Shore?\\
I've been catching waves with the dolphins all morning. I thought I would come and get some sunshine. What kind of victims do you expect to find in a tranquil place like this?\\
What do you know about that knight standing over there?\\
His armor is particularly garrish. You know I don't fraternize with land dwellers.\\
I don't know, I like when they're shiny like that. He looks like a giant fishing lure.\\
\textbf{Classified Utterance: } I suppose the only thing left to complete the illusion is for him to get wet.\\
\textbf{Correct Label: } Mermaid\\
\textbf{Prediction: } Sea Witch\\
\hline
\textbf{Context: }\_setting\_name Outside tower, Outside Tower\\
\_setting\_desc Moss grows from the tall stoic like structure adding to its mysterious presence. The stone walls appear insuperable like a mountain. The top is a pointed dome.\\
\_partner\_name enemy\\
\_self\_name horse\\
\_self\_persona We have hooves. four of them. and you can ride us. Oats please!\\
\textbf{Dialogue History:}\\
hello\\
hello there\\
\textbf{Classified Utterance:} What brings you here?\\
\textbf{Correct Label: } Horse\\
\textbf{Prediction: } Enemy\\
\hline
\textbf{Context:}
\textbf{Context:} \_setting\_name Royal Gardens, Outside Palace\\
\_setting\_desc Lined with rose bushes that look as if they have been watered by the God's, the Royal Gardens is a beauty to behold. An intricate labyrinth made of shrubs is at the center ending with a fountain. There are various benches on the sides of the rose bushes and a small lake in the back drop.\\
\_partner\_name king\\
\_self\_name a gardener pulling weeds\\
\_self\_persona I am the gardener of the castle. I plant thickets and plants. My work is beautiful.\\
\textbf{Dialogue History: }\\
Hi\\
\textbf{Classified Utterance:} Why hello there, your majesty!\\
\textbf{Correct Label:} a gardner\\
\textbf{Prediction:} king\\
\hline
\end{tabular}
\end{center}
\caption{Left-to-right dynamic classifier failure modes; see discussion in Section \ref{sec:rpa_failure_modes}.}
\label{tab:human_incorrect}
\end{table*}

\section{Human + Automatic Eval Correlation}
\label{sec:human_eval_correlation}
\begin{figure}
    \centering
    \includegraphics[width=\linewidth]{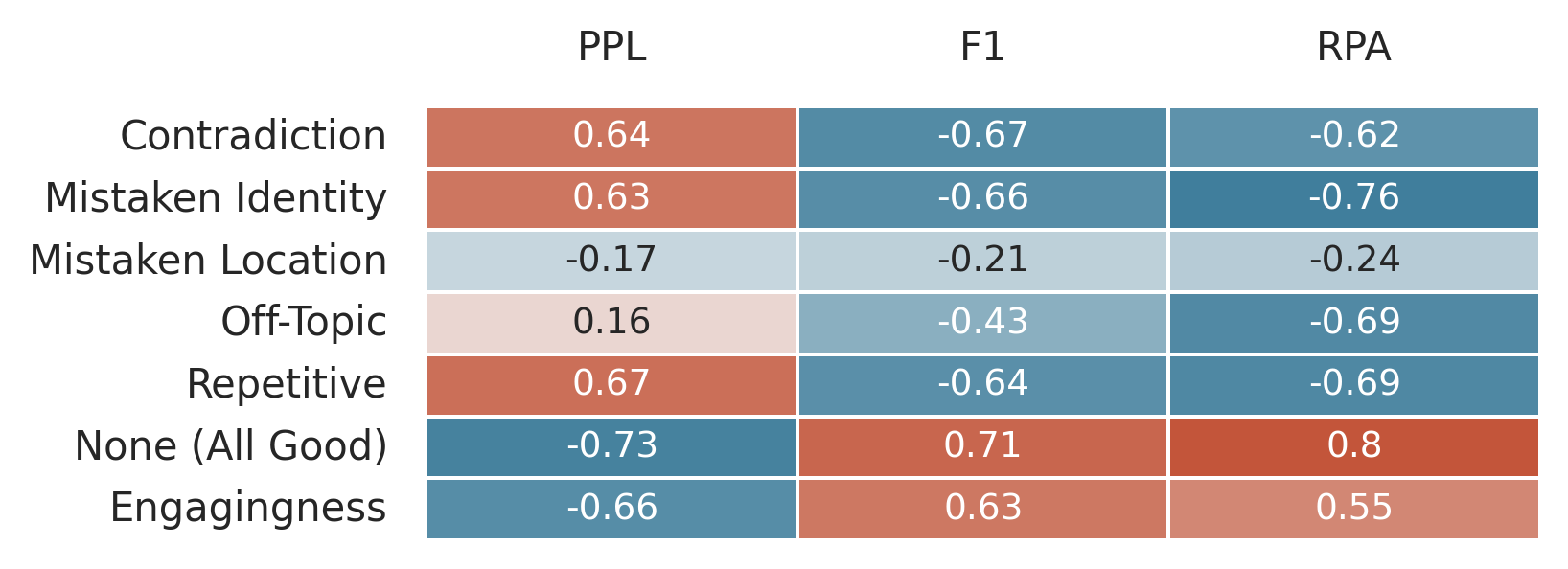}
    \caption{Correlation between human evaluations and automated metrics computed on the test set.}
    \label{fig:human_correlation}
\end{figure}
We analyze the correlation between human annotations and the automatic metrics collected on the LIGHT validation set, as shown in Figure \ref{fig:human_correlation}; we note some interesting trends:

\paragraph{Perplexity} perplexity appears to be positively correlated with mistaken identity, and negatively correlated with engagingness. So, perplexity is a good indicator of how fluent and engaging the model is in conversation, and can indirectly point to a better understanding of the role-playing task. An important note is that we only tested this amongst models of the same size, and only for the models we tested, so it is not clear that larger models will necessarily bring improvements.
\paragraph{F1} F1 word overlap is positively correlated with engagingness as well, so F1 may be a good proxy of model performance. Correlation with mistaken identity is negative here, implying that better F1 corresponds with better role-playing ability. However, we note that F1 is not a catch-all metric \citep{how_not_to_eval2016}.
\paragraph{RPA} RPA appears to be strongly negatively correlated with mistaken identity, indicating that it is indeed a good measure of the model's ability to stay in character. It is weakly negatively correlated with the other issues, and is somewhat positively correlated with engagingness as well. These correlations give us confidence that our RPA classifiers are adequately measuring role-playing ability within models.

\section{Per-Turn Analysis, Expanded}
\label{sec:per_turn_analysis_appendix}
\begin{figure*}
    \centering
    \includegraphics[width=0.6\linewidth]{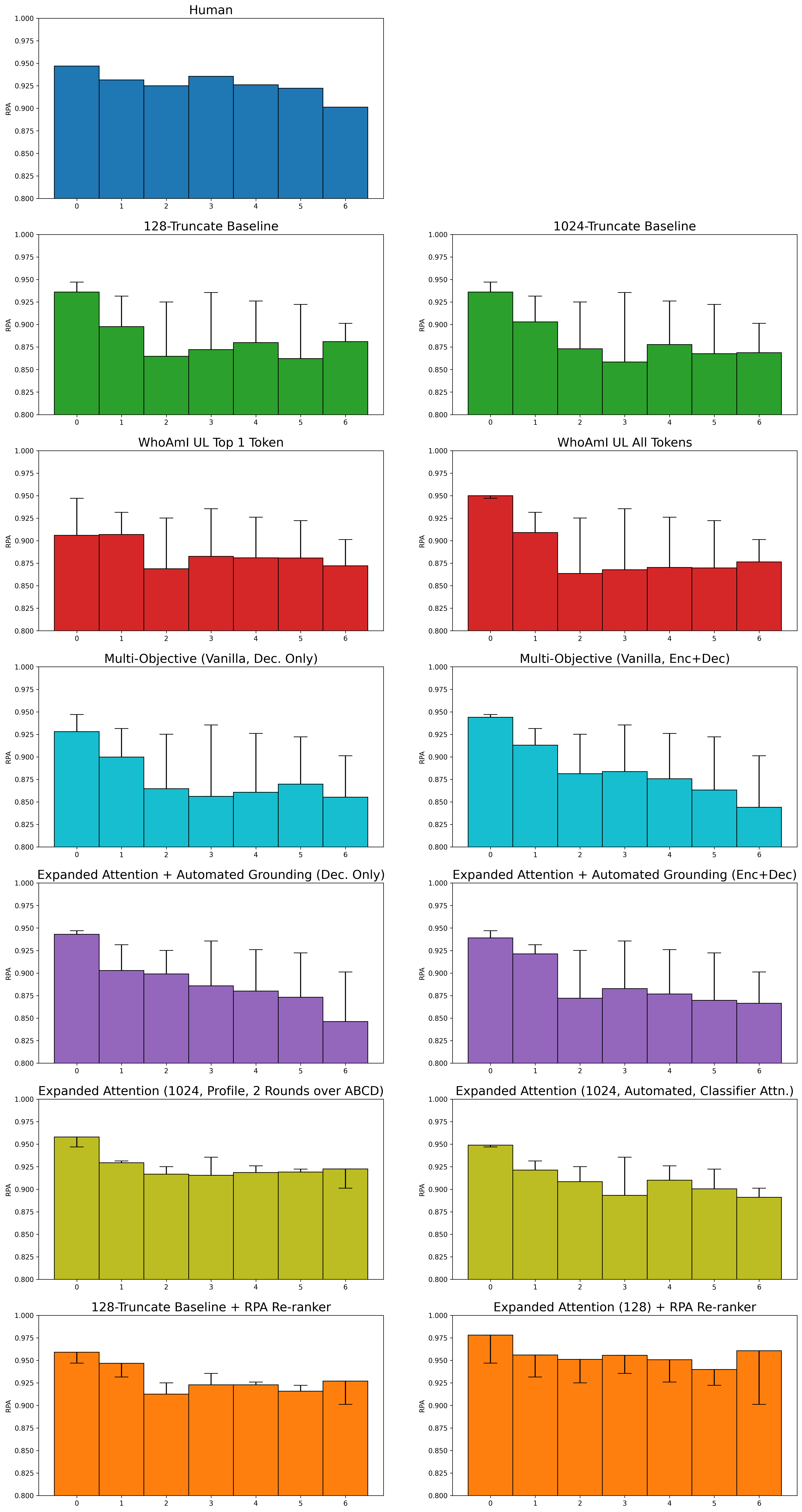}
    \caption{Per-turn RPA classifications, for a variety of models. Error bars show the difference between the model's RPA value and the human's RPA value.}
    \label{fig:per_turn_analysis_appendix}
\end{figure*}

In Figure \ref{fig:per_turn_analysis_appendix}, we see RPA results across turns of conversation for a wider variety of models.

\paragraph{Human} The human outputs are most often correct on the first turn, with gradual decay of accuracy throughout the conversation (according to RPA).
\paragraph{Vanilla \& Long Context} The vanilla baseline suffers a pretty dramatic drop off after the first couple of turns; the long-context model achieves slightly higher character accuracy overall but we see similar drop offs farther down the conversation.
\paragraph{RPA UL} The unlikelihood models seem to recover somewhat in the initial turns of conversation, however later turns still yield sharp drop offs in RPA.
\paragraph{Multi-objective} Similarly to the UL case, we see the most gains in initial turns compare to the vanilla baselines; however, we see even more dramatic drop offs towards the end of the conversation.
\paragraph{Expanded Attention} With profile grounding, we see near-human performance, with even better performance towards the end of the conversation. The automatic grounding improves over the baseline but is slightly worse than profile grounding. Combining automated grounding with multi-objective training leads to some benefits in earlier turns, but later turns still suffer.
\paragraph{Re-ranking} Although we're using the same RPA classifier to both re-ranker and score the model outputs, it is still interesting to examine on which turns the re-ranker benefits the model the most. We see in the last set of graphs that beam re-ranking seems to be most helpful in later turns, where other models generally drop off in efficacy.

\section{Expanded Attention Visualization}
\label{sec:heatmap_appendix}
\begin{figure*}
    \centering
    \includegraphics[height=1.25\linewidth]{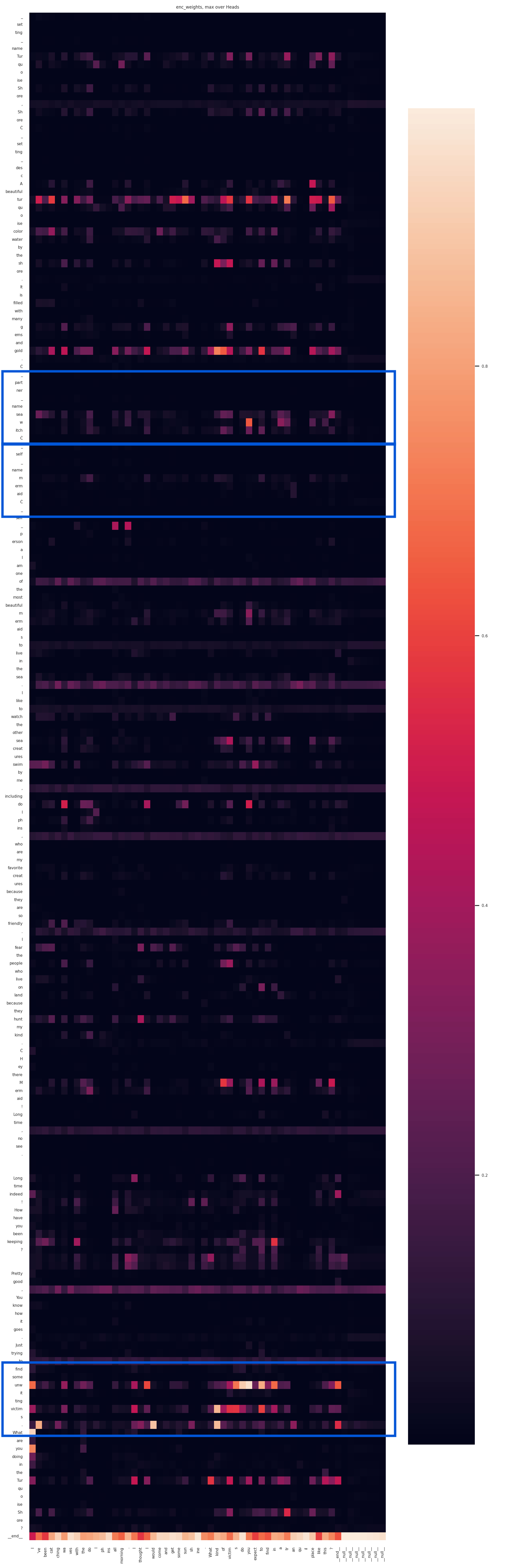}
    \caption{Vanilla Attention. The speaker here is the mermaid, whose partner is a sea-witch. The last utterance from the sea-witch is, ``What are you doing on the turquoise shore?''. The mermaid responds, ``I've been catching waves with the dolphins all morning. What kind of victims do you expect to find in a tranquil place like this?''. The vanilla model spreads its attention across the whole context; blue boxes at the top are attentions over the character descriptions, while the bottom box is attention over the word ``victims''.}
    \label{fig:no_extra_attention}
\end{figure*}

\begin{figure*}
    \centering
    \includegraphics[height=1.25\linewidth]{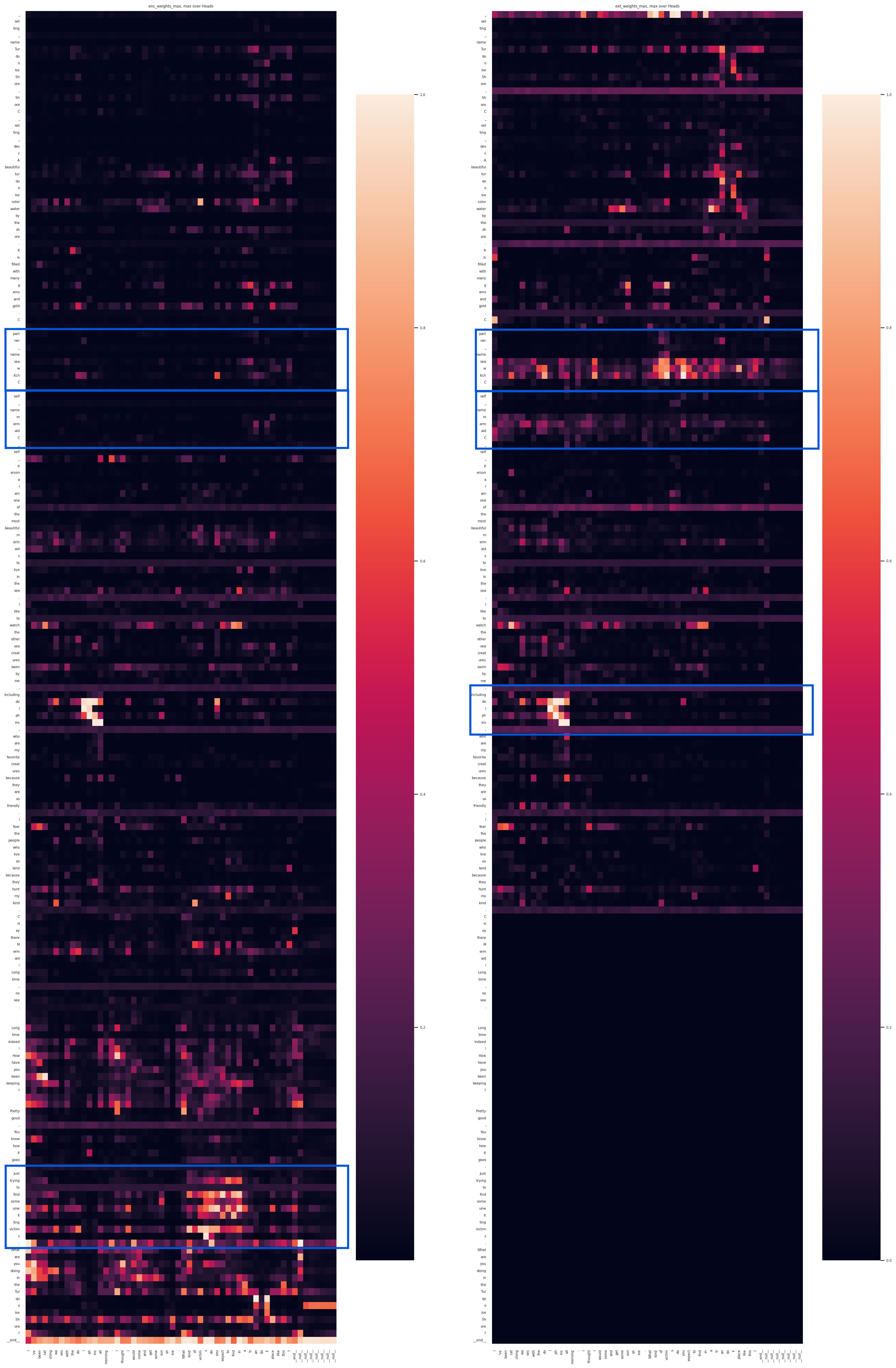}
    \caption{profile Expanded Attention. The speaker here is the mermaid, whose partner is a sea-witch. The last utterance from the sea-witch is, ``What are you doing on the turquoise shore?''. The mermaid responds, ``I've been catching waves with the dolphins all morning. What kind of victims do you expect to find in a tranquil place like this?''. \textbf{Left} original attention over the full context; \textbf{Right} expanded attention over the additional context. The top two boxes are the partner name and self name; the bottom box on the left refers to ``victims'', and on the right refers to the ``dolphins''.} 
    \label{fig:manual_attention}
\end{figure*}

To build the heat maps in Figures \ref{fig:no_extra_attention} and \ref{fig:manual_attention}, we look at the maximum attention applied per-head, and the maximum weight applied across the model decoder layers; other combinations were considered (mean per-head, mean over layers or last layer) and yielded similar findings.

The speaker is the mermaid, whose partner is a sea-witch. The last utterance from the sea-witch is, ``What are you doing on the turquoise shore?''. The mermaid responds, ``I've been catching waves with the dolphins all morning. What kind of victims do you expect to find in a tranquil place like this?''


\end{document}